\documentclass[journal]{IEEEtran}
\usepackage{cite}
\ifCLASSINFOpdf
\else
\fi
\usepackage[cmex10]{amsmath}
\usepackage{amssymb}
\usepackage{pgfplots}
\usepackage[ruled]{algorithm2e}
\usepackage{xcolor,colortbl}

\begin{document}
\title{Spectral Unmixing of Hyperspectral Imagery using Multilayer NMF}

\author{Roozbeh~Rajabi,~\IEEEmembership{Student~Member,~IEEE,}
        Hassan~Ghassemian,~\IEEEmembership{Senior~Member,~IEEE}% <-this % stops a space
%\thanks{Manuscript received Januray 20, 2014; revised March 10, 2014 and April 12, 2014.}
\thanks{The authors are with the Faculty of Electrical and Computer Engineering, Tarbiat Modares University, Tehran, Iran; e-mail: \{r.rajabi, ghassemi\}@modares.ac.ir.}} % <-this % stops a space

\markboth{IEEE GEOSCIENCE AND REMOTE SENSING LETTERS}%
{Shell \MakeLowercase{\textit{et al.}}: Bare Demo of IEEEtran.cls for Journals}



\maketitle

\begin{abstract}
Hyperspectral images contain mixed pixels due to low spatial resolution of hyperspectral sensors. Spectral unmixing problem refers to decomposing mixed pixels into a set of endmembers and abundance fractions. Due to nonnegativity constraint on abundance fractions, nonnegative matrix factorization (NMF) methods have been widely used for solving spectral unmixing problem. In this letter we proposed using multilayer NMF (MLNMF) for the purpose of hyperspectral unmixing. In this approach, spectral signature matrix can be modeled as a product of sparse matrices. In fact MLNMF decomposes the observation matrix iteratively in a number of layers. In each layer, we applied sparseness constraint on spectral signature matrix as well as on abundance fractions matrix. In this way signatures matrix can be sparsely decomposed despite the fact that it is not generally a sparse matrix. The proposed algorithm is applied on synthetic and real datasets. Synthetic data is generated based on endmembers from USGS spectral library. AVIRIS Cuprite dataset has been used as a real dataset for evaluation of proposed method. Results of experiments are quantified based on SAD and AAD measures. Results in comparison with previously proposed methods show that the multilayer approach can unmix data more effectively.
\end{abstract}


% \fi
\IEEEpeerreviewmaketitle

\section{Introduction}
\IEEEPARstart{D}{espite} high spectral resolution of hyperspectral sensors they have low spatial resolution. Low spatial resolution can cause mixed pixels in hyperspectral images. Mixed pixels contain more than one distinct material. These materials are called endmembers and the presence percentages of them in mixed pixels are called abundance fractions. Spectral unmixing problem refers to decomposing the measured spectra of mixed pixels into a set of endmembers and their abundance fractions \cite{MLNMF:SPM02_Keshava}.

Linear mixing model is often used for solving spectral unmixing problem because of its simplicity and efficiency in most cases \cite{MLNMF:SPM02_Keshava}. There are many methods proposed for solving spectral unmixing problem \cite{TGRS05_VCA,SPIE09_N-FINDR,IGRASS08_MVSA,MLNMF:Springer11_Baysian,TGRS12_DECA,TGRS11_Sparse}. They can be categorized in geometrical, statistical and sparse regression based approaches \cite{MLNMF:JSTARS12_Overview}. Due to nonnegativity constraint in linear mixing model, nonnegative matrix factorization (NMF) has been widely used for solving spectral unmixing problem \cite{TGRS07_MVC-NMF,TIP11_SparseNMF,MLNMF:TGRS11_L05NMF}. Generally, algorithms based on NMF lead to a NP-hard optimization problem \cite{MLNMF:NP-Hard,SPM14_Perspective}. In this letter, an algorithm called MLNMF method, based on multilayer NMF \cite{EL06_MLNMF} has been proposed to improve the performance of NMF methods for hyperspectral data unmixing. Using multilayer structure of MLNMF, spectral signatures matrix is considered as a product of sparse matrices. In each layer, sparseness constraint on endmembers matrix is added to the cost function. To evaluate the proposed method, synthetic and real datasets are used. Synthetic data is generated using USGS spectral library and synthetic images. AVIRIS Cuprite Nevada dataset is used as a real data to examine the proposed algorithm. Spectral unmixing results are evaluated based on spectral angle distance (SAD) and abundance angle distance (AAD) metrics and compared against vertex component analysis (VCA) \cite{TGRS05_VCA} and $L_{1/2}$-NMF \cite{MLNMF:TGRS11_L05NMF}.

The rest of this letter is organized as follows. Problem definition and methodology framework are discussed in section II. Section III presents metrics that are used for the evaluation of the proposed method. Experiments and results using synthetic and real datasets are also summarized in this section. Finally, section IV discusses the results and concludes the paper.

\section{methodology}
There are two main classes of mixing models for spectral unmixing problem: linear mixing model (LMM) and the category of nonlinear mixing models \cite{SPM14_Nonlinear}. LMM is not always true but it is an acceptable model in many scenarios and is widely used to solve spectral unmixing problem \cite{SPM14_Perspective}. In this letter only LMM is considered. Mathematical formulation of LMM, sparse NMF and proposed MLNMF method are described in this section.

%\hfill December 27, 2012

\subsection{Linear Mixing Model (LMM)}
Mathematical formulation of linear mixture model is expressed in (\ref{lmm}).
\begin{equation}
\label{lmm}
\mathbf{X = AS+E}.
\end{equation}

In this equation $\mathbf{X}\in{\mathfrak{R}^{B\times{N}}}$  refers to the observation matrix, $B$ denotes the number of spectral bands and $N$ denotes the total number of pixels. $\mathbf{A}\in{\mathfrak{R}^{B\times{P}}}$ and $\mathbf{S}\in{\mathfrak{R}^{P\times{N}}}$ refer to the signatures and the abundance fractions matrices respectively. $P$ denotes the number of endmembers and  $\mathbf{E}\in{\mathfrak{R}^{B\times{N}}}$  refers to observation noise.

This model is subject to two physical constraints on the abundance fraction values: abundance nonnegativity constraint (ANC) and abundance sum to one constraint (ASC) \cite{MLNMF:SPM02_Keshava}. These constraints are formulated in the following equations:
\begin{equation}
\label{anc}
\forall{i,j}:{s}_{ij}\geq{0},
\end{equation}
\begin{equation}
\label{asc}
\sum_{i=1}^P{{s}_{ij}=1}.
\end{equation}
%\IEEEpubidadjcol

%\subsubsection{Subsubsection Heading Here}
%Subsubsection text here.

\subsection{Sparse Single Layer Nonnegative Matrix Factorization}
NMF is an efficient method for decomposing multivariate data. Algorithms based on different cost functions can be used to solve NMF problem \cite{ANIPS00_AlgorithmsNMF}. In this paper, Euclidean distance is used. General NMF problem is formulated in (\ref{nmf}):
\begin{equation}
\label{nmf}
\mathbf{X\approx{AS}}.
\end{equation}

In this equation $\mathbf{X}$ is a $B\times{N}$ nonnegative matrix, $\mathbf{A}$ is a $B\times{P}$  nonnegative matrix and  $\mathbf{S}$ is a $P\times{N}$  nonnegative matrix. The cost function in (\ref{onmf}) can be used for solving NMF problem. This cost function should be minimized with respect to  $\mathbf{A}$ and $\mathbf{S}$ subject to the nonnegativity constraints on $\mathbf{A}$ and $\mathbf{S}$. Multiplicative update algorithm is an efficient way of minimizing the cost function in NMF problem, since it is fast and easy to implement \cite{ANIPS00_AlgorithmsNMF}.
\begin{equation}
\label{onmf}
\mathcal{O}_{\text{NMF}}=\tfrac{1}{2}\lVert{\mathbf{X-AS}}\rVert_{F}^{2}.
\end{equation}

In hyperspectral unmixing, one of the constraints that can be used is a sparseness constraint on the abundance fractions matrix. The cost function using this constraint is given in (\ref{oslnmf}). From physical point of view, the logic for this constraint is that the number of endmembers present in each mixed pixel is much less than the total number of endmembers.
\begin{equation}
\label{oslnmf}
\mathcal{O}_{\text{SLNMF}}=\tfrac{1}{2}\lVert{\mathbf{X-AS}}\rVert_{F}^{2}+\alpha\lVert{\mathbf{S}}\rVert_{1/2},
\end{equation}
where $\alpha$ is the regularization parameter that controls the impact of sparseness constraint [12]. $q$-norm of a matrix $\mathbf{S}$ is defined in (\ref{qnorm}).
\begin{equation}
\label{qnorm}
\lVert{\mathbf{S}}\rVert_{q}=\sum_{i=1}^P{\sum_{j=1}^N{{s}_{ij}^q}},
\end{equation}
where ${s}_{ij}$ is an element of matrix $\mathbf{S}$ in the $i^{th}$ row and $j^{th}$ column.

To take care of ASC constraint on the abundance fraction values, fully constrained least squares (FCLS) method \cite{TGRS01_FCLS} has been used. In this method new observation and signature matrices are defined as:

\begin{equation}
\label{fcls}
\widetilde{\mathbf{X}}=\begin{bmatrix}
\mathbf{X}\\
\delta\mathbf{1}
\end{bmatrix},
\widetilde{\mathbf{A}}=\begin{bmatrix}
\mathbf{A}\\
\delta\mathbf{1}
\end{bmatrix},
\end{equation}
where $\delta$ is a parameter that controls the impact of ASC constraint and $\mathbf{1}$ is a row vector with all elements equal to one \cite{TGRS01_FCLS}.

\subsection{Proposed MLNMF Method}
\begin{figure}[!t]
   \centering
   \includegraphics{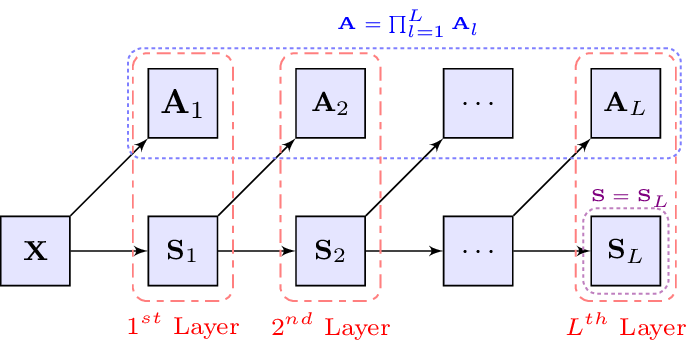}
   \newlength\figureheight
   \newlength\figurewidth
   \caption{Multilayer NMF decomposition.}
   \label{fig_multilayer}
   \end{figure}

In this letter, in order to improve the performance of NMF methods for hyperspectral unmixing, using MLNMF method is proposed. MLNMF has been initially proposed in \cite{EL06_MLNMF} for solving blind source separation problem in signal processing. In this method, multilayer structure is used to decompose the observation matrix. In the first layer, basic decomposition is done resulting in $\mathbf{A}_{1}$ and $\mathbf{S}_{1}$. Then in the second layer, the result of the first layer ($\mathbf{S}_{1}$) is decomposed into $\mathbf{A}_{2}$ and $\mathbf{S}_{2}$. This process will be repeated to reach the maximum number of layers ($L$) (see \figurename{~\ref{fig_multilayer}}). The mathematical representation of multilayer structure is summarized in (\ref{multilayer}).
\begin{equation}
\begin{aligned}
\label{multilayer}
&\mathbf{X}=\mathbf{A}_1\mathbf{S}_1,\;
\mathbf{S}_1=\mathbf{A}_2\mathbf{S}_2,\;
\hdots,\;
\mathbf{S}_{L-1}=\mathbf{A}_L\mathbf{S}_L,&\\
&\Rightarrow\;\mathbf{A}=\mathbf{A}_1\mathbf{A}_2\ldots\mathbf{A}_L,\;\;\;\mathbf{S}=\mathbf{S}_L.&
\end{aligned}
\end{equation}

In the proposed method, we used the sparsity constraints for both spectral signatures and abundance fractions. Applying sparsity constraint on abundance fractions is already well discussed in \cite{MLNMF:TGRS11_L05NMF}. In this letter, we also added sparsity constraint on the spectral signatures matrix in each layer. Although the true spectral signatures matrix is not sparse, the decomposition result of it in each layer can be sparse. The intuitive for this assumption is that we can represent a nonsparse matrix as a product of a few sparse matrices. The mathematical proof of this assumption is still an open problem \cite{ANCA07_MLNMF}. Note that in MLNMF method $\mathbf{X}$ is decomposed partially in the first layer and will be decomposed completely in a number of layers. Although in the first layer $\mathbf{A}$ and $\mathbf{S}$ are obtained but in the succeeding layers the results will be improved. Considering the mentioned constraints, MLNMF cost function for the $l^{th}$ layer is defined in (\ref{omlnmf}).
\begin{equation}
\label{omlnmf}
\mathcal{O}_{\text{MLNMF}}=\tfrac{1}{2}\lVert{\mathbf{X}_l-\mathbf{A}_l\mathbf{S}_l}\rVert_{F}^{2}+\alpha_A\lVert{\mathbf{A}_l}\rVert_{1/2}+\alpha_S\lVert{\mathbf{S}_l}\rVert_{1/2},
\end{equation}

In our experiments, $\alpha_A$  is set using the following equation:
\begin{equation}
\label{alpha}
\alpha_A=\alpha_{0}e^{\frac{-t}{\tau}},
\end{equation}
where $t$ is the iteration number in the process of optimization and, $\alpha_{0}$  and $\tau$  are constants to regularize the impact of sparsity constraints. This equation for regularization parameters is motivated by temperature function in the simulated annealing (SA) method and can avoid getting stuck in local minima \cite{EL06_MLNMF}. Also, to increase the impact of sparsity constraint on the abundance fractions, $\alpha_S$ could be chosen larger than $\alpha_A$. As a rule of thumb, we set $\alpha_S=2\alpha_A$ in our experiments, though one can optimize the algorithm by finding the best relationship between these parameters.

By differentiating (\ref{omlnmf}) with respect to $\mathbf{A}_l$  and $\mathbf{S}_l$, multiplicative update rules can be calculated. These rules for solving the NMF problem with $L_1$ sparseness constraints are calculated in \cite{LAA06_NMFforSpectral}. We have obtained the update rules by substituting the terms related to $L_{1/2}$ constraints \cite{MLNMF:TGRS11_L05NMF}. The resulted multiplicative update rules for our NMF problem are expressed in (\ref{updatea}) and (\ref{updates}).
\begin{equation}
\label{updatea}
\mathbf{A}_l\leftarrow\mathbf{A}_l.*(\mathbf{X}_l\mathbf{S}_l^\mathbf{T})./(\mathbf{A}_l\mathbf{S}_l\mathbf{S}_l^\mathbf{T}+\frac{1}{2}\alpha_{A}\mathbf{A}_l^{-1/2}),
\end{equation}
\begin{equation}
\label{updates}
\mathbf{S}_l\leftarrow\mathbf{S}_l.*(\mathbf{A}_l^\mathbf{T}\mathbf{X}_l)./(\mathbf{A}_l^\mathbf{T}\mathbf{A}_l\mathbf{S}_l+\frac{1}{2}\alpha_{S}\mathbf{S}_l^{-1/2}),
\end{equation}
where $(.)^{\mathbf{T}}$ denotes the transpose of a matrix.

ASC constraint has been considered using (\ref{fcls}). After evaluating performance of the algorithm for different values of $\delta$, we set $\delta=25$ in our experiments.

To initialize the algorithm, results of VCA \cite{TGRS05_VCA} have been used in the experiments. Note that random initialization can also be used, but VCA initialization gives more robust results. Another important setting of the algorithm is the stopping criteria. In each layer, the algorithm will be stopped either after reaching the maximum number of iterations ($T_{max}$) or meeting the stopping criteria in (\ref{stop}) for ten successive iterations.
\begin{equation}
\label{stop}
\lVert{\mathcal{O}_{\text{MLNMF}_{\text{New}}}-\mathcal{O}_{\text{MLNMF}_{\text{Old}}}}\rVert<\epsilon,
\end{equation}
where $\mathcal{O}_{\text{MLNMF}_{\text{New}}}$ and $\mathcal{O}_{\text{MLNMF}_{\text{Old}}}$  are cost function values for iterations $t$ and $t-1$ respectively and $\epsilon$ is the error value that has been set to $10^{-4}$ in our experiments. The proposed MLNMF algorithm is summarized in Algorithm \ref{MLNMF}.

%\begin{algorithm}
%\caption{MLNMF Algorithm for hyperspectral unmixing}
%\begin{algorithmic}
%\FOR{$l=1$ \TO $L_{max}$ } 
%\STATE{
%initialize $\mathbf{A}$ and $\mathbf{S}$ using VCA for the first layer and using random initialization for other layers
%\FOR{$l=1$ \TO $L_{max}$ } 
%\STATE{
%update $\mathbf{A}$ using (9)
%$\mathbf{X}=$ , $\mathbf{A}=$ using (11)
%update $\mathbf{S}$ using (10)
%\IF{stopping criteria in (12)} \STATE{break} \ENDIF
%}\ENDFOR
%} \ENDFOR
%\STATE do something 
%\end{algorithmic}
%\end{algorithm}

\begin{algorithm}
\label{MLNMF}
\SetKwData{Left}{left}
\SetKwData{This}{this}
\SetKwData{Up}{up}
\SetKwFunction{Union}{Union}
\SetKwFunction{FindCompress}{FindCompress}
\SetKwInOut{Input}{input}
\SetKwInOut{Output}{output}
\caption{MLNMF Algorithm}
\Input{Observation matrix ($\mathbf{X}$)\\
%Initial values for $\mathbf{A}$ and $\mathbf{S}$\\
Parameters: $P$, $\alpha_0$ , $\tau$ , $\delta$ , $L$ and $T_{max}$
}
\Output{Estimated spectral signatures and abundance fractions ($\mathbf{A}$ and $\mathbf{S}$)}
$\mathbf{X}_1=\mathbf{X}$\;
\For{$l=1$ \KwTo $L$}{
Initialize $\mathbf{A}_l$ and $\mathbf{S}_l$ using VCA for the first layer and using random initialization for other layers\\
\For{$t=1$ \KwTo $T_{max}$}{
update $\mathbf{A}_l$ using (\ref{updatea})\;
$\mathbf{X}_l=\widetilde{\mathbf{X}}_l$ , $\mathbf{A}_l=\widetilde{\mathbf{A}}_l$ using (\ref{fcls})\;
update $\mathbf{S}_l$ using (\ref{updates})\;
\If{stopping criteria in (\ref{stop})}{break}
}
$\mathbf{X}_{l+1}=\mathbf{S}_l$
}
$\mathbf{A}=\prod_{l=1}^{L}\mathbf{A}_l$ and $\mathbf{S}=\mathbf{S}_{L}$
\end{algorithm}

\section{Experiments and Results}
For evaluation purposes, synthetic data and real dataset have been used in this research. In subsection A, evaluation metrics used for quantifying results have been introduced. Experiments for evaluation of the proposed scheme are described and results are shown for synthetic and real dataset in subsections B and C respectively.

\subsection{Evaluation Measures}
For evaluation of the proposed method, two different measures have been used: spectral angle distance (SAD) and abundance angle distance (AAD) \cite{TIP07_MaximumEntropy}. SAD measures the similarity between original spectral signatures ($\mathbf{m}_{i}$) and estimated ones ($\hat{\mathbf{m}}_{i}$) as formulated in (\ref{sad}).
\begin{equation}
\label{sad}
\text{SAD}_{\mathbf{m}_{i}}=\cos^{-1}{(\frac{\mathbf{m}_{i}^{\mathbf{T}}\hat{\mathbf{m}}_{i}}{\lVert{\mathbf{m}_{i}}\rVert\lVert{\hat{\mathbf{m}}_{i}}\rVert})}
\end{equation}
AAD measures the similarity between original abundance fractions ($\mathbf{a}_{i}$) and estimated ones ($\hat{\mathbf{a}}_{i}$) as formulated in (\ref{aad}).
\begin{equation}
\label{aad}
\text{AAD}_{\mathbf{a}_{i}}=\cos^{-1}{(\frac{\mathbf{a}_{i}^{\mathbf{T}}\hat{\mathbf{a}}_{i}}{\lVert{\mathbf{a}_{i}}\rVert\lVert{\hat{\mathbf{a}}_{i}}\rVert})}
\end{equation}

Advantage of using these measures is that they are independent of scale difference between original and estimated vectors. To have an overall measure of performance, root mean square of these measures are defined in (\ref{rmssad}) and (\ref{rmsaad}).
\begin{equation}
\label{rmssad}
\text{rmsSAD}=(\frac{1}{P}\sum\limits_{i=1}^{P}({\text{SAD}_{\mathbf{m}_{i}}})^{2})^{1/2}
\end{equation}
\begin{equation}
\label{rmsaad}
\text{rmsAAD}=(\frac{1}{N}\sum\limits_{i=1}^{N}({\text{AAD}_{\mathbf{a}_{i}}})^{2})^{1/2}
\end{equation}

\subsection{Experiment I: (synthetic data) USGS Spectral Library}
\begin{figure}[!t]
      \includegraphics{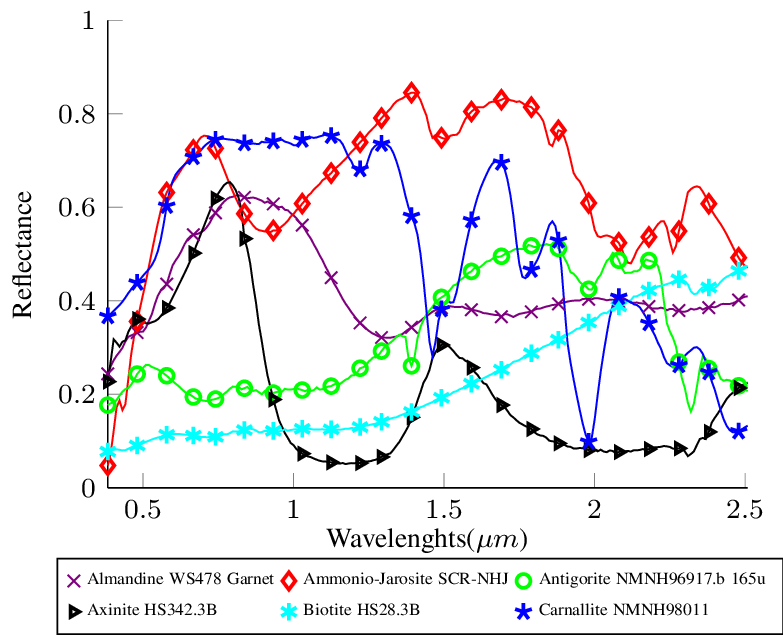}
    \caption{Selected Materials from USGS Library (splib06).}
    \label{fig_SelectedMats}
   \end{figure}

Spectral signatures from USGS library (splib06) \cite{USGS07_splib06} have been used in this paper to generate simulated data. Spectral signatures in this library contain 224 spectral bands with the spectral resolution of $10nm$ that covers wavelength range of $380nm$ to $2500nm$. Six signatures of this library have been chosen randomly and shown in \figurename{~\ref{fig_SelectedMats}}. Then the procedure described in \cite{TGRS07_MVC-NMF} has been used to create synthetic images of size $64\times64$ pixels containing no pure pixels. To do so, each image is divided into $8\times8$ blocks and all pixels in each block are filled up by one of signatures randomly selected from the chosen signatures. To create linear mixture, a spatial low pass filter of size $9\times9$ has been applied to the image. To remove probable pure pixels in the resulted image, all pixels with abundances greater than $80\%$ have been replaced by a mixture of all endmembers with equally distributed abundances. Finally, in order to simulate sensor noise and possible measurement errors, zero mean Gaussian noise is added to the data.

%   \begin{figure}[!t]
%   \centering
%   \includegraphics[width=3in]{SelectedMats}
%   \caption{Band 30 of the subimage of AVIRIS Cuprite Nevada Dataset.}
%   \label{fig_SelectedMats}
%   \end{figure}

In this experiment, robustness and performance of the algorithm in the presence of noise are investigated. Synthetic dataset is used because when using this kind of data, true spectral signatures and abundance fractions are completely known. Zero mean Gaussian noise with different levels of SNR has been added to data to simulate the measurement error and sensor noise \cite{TGRS07_MVC-NMF}. The relationship between SNR and zero mean noise variance is given in the following equations:
\begin{equation}
\begin{aligned}
\label{snr}
&\text{SNR}=10\log_{10}(E[\mathbf{x}^{\mathbf{T}}\mathbf{x}]/E[\mathbf{e}^{\mathbf{T}}\mathbf{e}]),&\\	
&\sigma_{\mathbf{e}}^2=E[\mathbf{x}^{\mathbf{T}}\mathbf{x}]/(10^{\text{SNR}/10}),&
\end{aligned}
\end{equation}
where $\mathbf{x}$  and $\mathbf{e}$ represent signal and noise levels of a pixel respectively and $E[.]$ denotes the expectation operator.

\begin{figure}[!t]
   \centering
%   \hspace{-20pt}
   \includegraphics{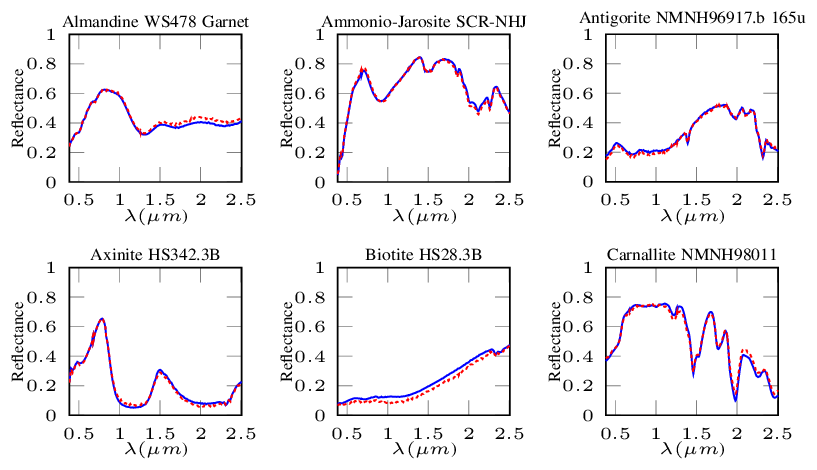}
   \caption{Original spectral signatures (blue solid lines) and estimated ones using MLNMF (red dashed lines)}
   \label{fig_syntheticExtracted}
   \end{figure}

Parameters of the algorithm are selected as follows: $\alpha_{0}=0.1$, $\tau=25$, $L=10$ and $T_{max}=400$. These parameters are determined to gain the best results but because of the lack of space we did not present the parameter selection procedure here. Estimated spectral signatures and the original ones are depicted in \figurename{~\ref{fig_syntheticExtracted}}. In this experiment SNR has been set to $20dB$.

\figurename{~\ref{fig_rmsSADrmsAAD} shows the performance of the proposed method in comparison with $L_{1/2}$-NMF \cite{MLNMF:TGRS11_L05NMF} and VCA \cite{TGRS05_VCA} methods. Since VCA can only extract endmembers, FCLS \cite{TGRS01_FCLS} has been used in conjunction with VCA to extract endmembers. \figurename{~\ref{fig_meanvariance} compares the stability of the results in terms of mean and standard deviation for 20 runs of the algorithms. As it can be seen in these figures, our method excels $L_{1/2}$-NMF and VCA.

   \begin{figure}[!t]
   \centering
   \includegraphics{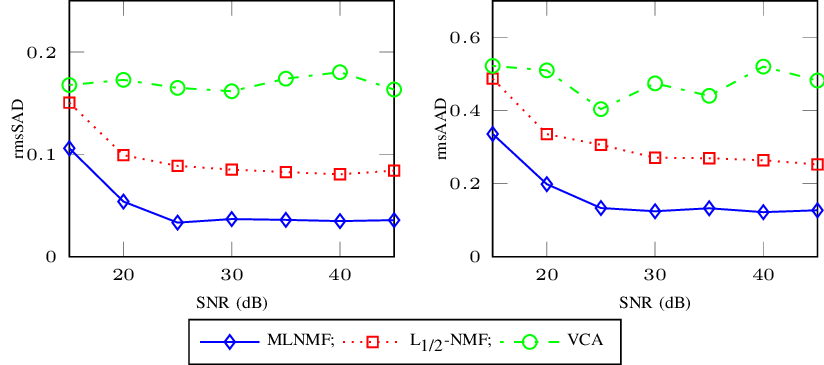}
   \caption{Comparison of rmsSAD (left) and rmsAAD (right) for VCA,  $L_{1/2}$-NMF and MLNMF vs. SNR.}
   \label{fig_rmsSADrmsAAD}
   \end{figure}
   
      \begin{figure}[!t]
      \centering
         \includegraphics{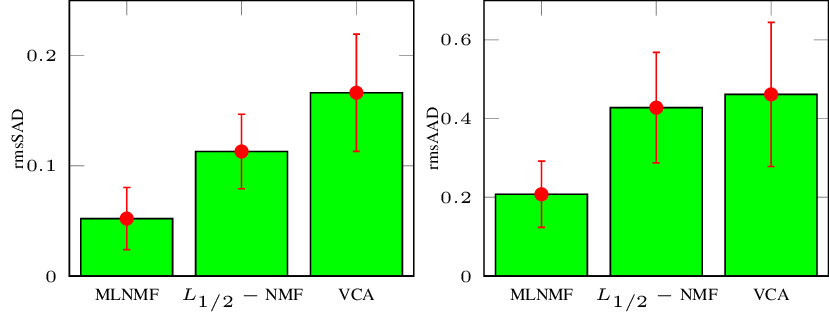}
      \caption{Comparison of methods in terms of mean and standard deviation of rmsSAD (left) and rmsAAD (right) for 20 Runs.}
      \label{fig_meanvariance}
      \end{figure}

\subsection{Experiment II: (real data) AVIRIS Cuprite Nevada}
The Cuprite, Nevada dataset collected by AVIRIS sensor is used in this experiment \cite{MLNMF:AVIRIS_Cuprite}. In this letter, a $250\times191$-pixel subscene of this dataset has been used similar to \cite{TGRS05_VCA}. Band 30 of this subscene is illustrated in \figurename{~\ref{fig_cuprite30}}. After removing low SNR and water absorption bands, a total number of 188 bands remained and used in the experiment.

   \begin{figure}[!t]
   \centering
   \includegraphics[width=0.9in]{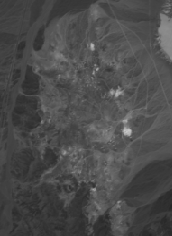}
   \caption{Band 30 of the subimage of AVIRIS Cuprite Nevada Dataset}
   \label{fig_cuprite30}
   \end{figure}

Based on the study in \cite{MLNMF:MineralMap_Cuprite} and results in \cite{TGRS05_VCA}} there are 14 materials in the scene. But some of them are very similar and we considered just one of each similar pair. Also we added ''Chalcedony'' as one of the endmembers according to the mineral map in \cite{MLNMF:MineralMap_Cuprite}. In this way, 12 endmembers are selected as true signatures.

In this experiment the regularization parameters are set to $\alpha_{0}=0.1$  and $\tau=25$. Estimated signatures along with the true signatures of USGS AVIRIS convolved spectral library (s06av95a) \cite{USGS07_splib06} are demonstrated in \figurename{~\ref{fig_cupritesignatures}}. Extracted abundance fraction maps using MLNMF method are illustrated in \figurename{~\ref{fig_cupriteabundances}}. Table I summarizes the results of the experiment on Cuprite dataset. As the results show, the proposed method outperforms VCA and $L_{1/2}$-NMF methods in terms of rmsSAD.

   \begin{figure}[!t]
   \centering
	\includegraphics{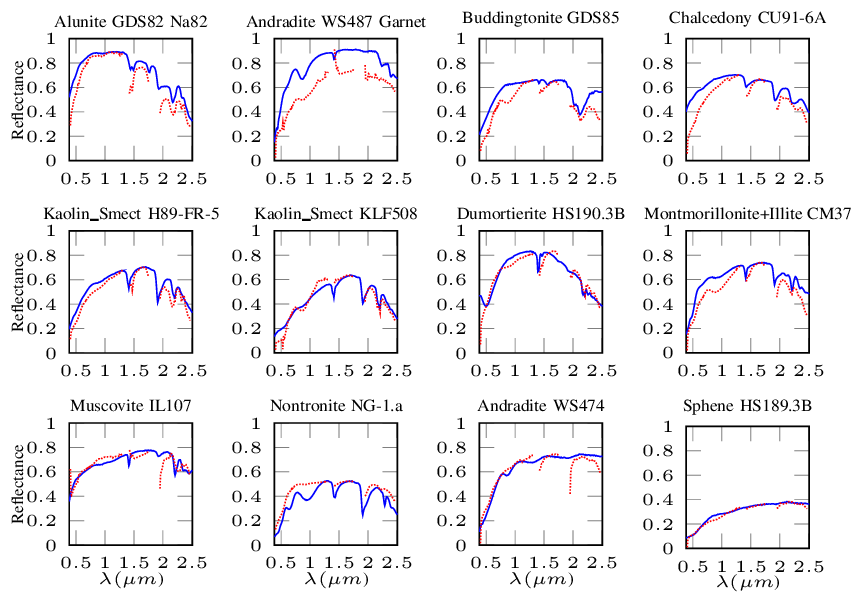}
   \caption{Original spectral signatures (blue solid lines) and estimated ones using MLNMF (red dotted lines)}
   \label{fig_cupritesignatures}
   \end{figure}

   \begin{figure}[!t]
   \centering
   \includegraphics{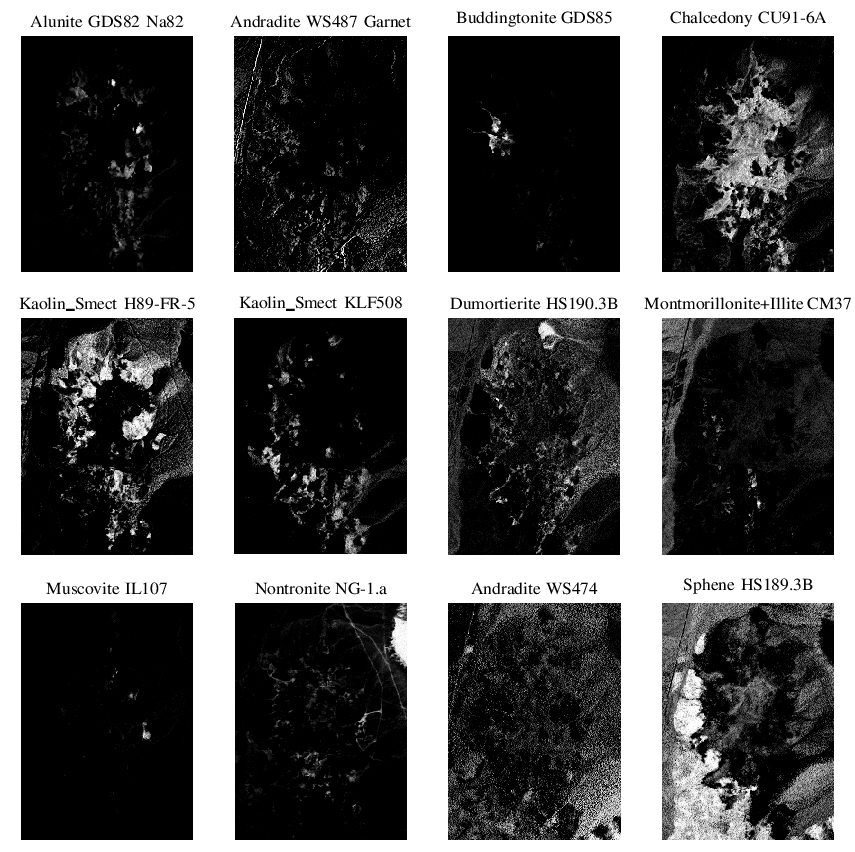}
   \caption{Estimated abundance fractions maps using MLNMF.}
   \label{fig_cupriteabundances}
   \end{figure}

\begin{table}[!t]
\renewcommand{\arraystretch}{1.3}
% if using array.sty, it might be a good idea to tweak the value of
% \extrarowheight as needed to properly center the text within the cells
\caption{Comparison between methods in terms of SAD}
\label{cupriteresultstable}
\centering
% Some packages, such as MDW tools, offer better commands for making tables
% than the plain LaTeX2e tabular which is used here.
\begin{tabular}{cccc}
%\cline{2-4}
 &  & Methods\\
\cline{2-4}
Material Name & VCA & $L_{1/2}$-NMF & MLNMF\\
\hline \hline
Alunite GDS82 Na82 & 0.1741	& \textbf{0.0655}  & 0.0948\\
\hline
Andradite WS487 Garnet &	0.1445&	\textbf{0.0721} & 0.1045\\
\hline
Buddingtonite GDS85 D-206&	0.1143&	\textbf{0.1104} & 0.1319\\
\hline
Chalcedony CU91-6A&	\textbf{0.1168}&	0.1327&	0.1387\\
\hline
Kaolin\_Smect H89-FR-5 .3Kaol&	0.0585&	\textbf{0.0448}&	0.0775\\
\hline
Kaolin\_Smect KLF508 .85Kaol&	0.0856&	\textbf{0.0847}&	0.0981\\
\hline
Dumortierite HS190.3B&	0.1096&	0.0968&	\textbf{0.0856}\\
\hline
Montmorillonite+Illite CM37&	0.0740&	\textbf{0.0429}	&0.0921\\
\hline
Muscovite IL107	& 0.0845&	0.1604&	\textbf{0.0745}\\
\hline
Nontronite NG-1.a&	\textbf{0.0717}&	0.0817&	0.1177\\
\hline
Andradite WS474	&0.0996&	0.0982&	\textbf{0.0739}\\
\hline
Sphene HS189.3B	&0.0561&	0.2171&	\textbf{0.0508}\\
\hline \hline
rmsSAD&	0.1047	& 0.1138 &	\textbf{0.0981}\\
\hline
\end{tabular}
\end{table}

\section{Conclusion}
Methods based on NMF are one of the promising methods for hyperspectral unmixing purposes. In this letter we proposed using multilayer nonnegative matrix factorization for hyperspectral unmixing to improve the results and reduce the risk of getting stuck in local minima. We considered sparsity constraint for both spectral signatures and abundance fractions. From physical point of view, although spectral signatures matrix is not sparse, but by considering multilayer structure, we can decompose it into a few sparse matrices. The proposed method has been applied on synthetic and real datasets. Results are presented in terms of SAD and AAD metrics and compared against VCA and $L_{1/2}$-NMF. Comparisons show that the proposed algorithm can unmix hyperspectral data more effectively.

% or


%\appendices

%\section{}

\section*{Acknowledgment}
The authors would like to thank the anonymous reviewers for their valuable and helpful comments and suggestions.

\ifCLASSOPTIONcaptionsoff
  \newpage
\fi





\bibliographystyle{IEEEtran}

%\begin{thebibliography}{1}
\bibliography{IEEEabrv,MLNMF}
%\end{thebibliography}


%\end{IEEEbiography}


%\newpage



%\vfill


\end{document}